\documentclass[12pt]
{llncs}

\usepackage[utf8]{inputenc}
\usepackage[T1]{fontenc}
\usepackage{graphicx}
\usepackage{url}

\begin{document}

\title{Readable Twins of Unreadable Models}

\author{Krzysztof Pancerz\inst{1,5}\orcidID{0000-0002-5452-6310} \and Piotr Kulicki\inst{1}\orcidID{0000-0001-5413-3886} \and Micha{\l} Kalisz\inst{1}\orcidID{0000-0003-1955-1884} \and Andrzej Burda\inst{1,2}\orcidID{0000-0002-1962-0264} \and Maciej Stanis{\l}awski\inst{3,5}\orcidID{0009-0002-9586-6597} \and Jaromir Sarzy\'{n}ski\inst{4}\orcidID{0000-0002-2133-4187}}

\institute{The John Paul II Catholic University of Lublin, Poland \\
\texttt{\{kpancerz, kulicki, mkalisz\}@kul.pl} \and
University of Economics and Human Sciences in Warsaw, Poland\\
\texttt{a.burda@vizja.pl} \and
University of Warmia and Mazury, Olsztyn, Poland \\
\texttt{maciej.stanislawski@uwm.edu.pl}
\and
University of Rzeszów, Poland\\
\texttt{jsarzynski@ur.edu.pl}
\and MakoLab S.A., {\L}\'{o}dz, Poland 
}

\date{}

\maketitle

\begin{abstract}
Creating responsible artificial intelligence (AI) systems is an important issue in contemporary research and development of works on AI. One of the characteristics of responsible AI systems is their explainability. In the paper, we are interested in explainable deep learning (XDL) systems. On the basis of the creation of digital twins of physical objects, we introduce the idea of creating readable twins (in the form of imprecise information flow models) for unreadable deep learning models. The complete procedure for switching from the deep learning model (DLM) to the imprecise information flow model (IIFM) is presented. The proposed approach is illustrated with an example of a deep learning classification model for image recognition of handwritten digits from the MNIST data set.
\end{abstract}

\section{Introduction}

One of the most current problems in Artificial Intelligence (AI) is to make AI tools human-readable (interpretable, explainable, etc.) and in consequence to make AI responsible (cf. \cite{BARREDOARRIETA202082}). AI tools are currently, in many cases, reinforced by deep neural networks (DNNs). Therefore, special attention in research on Explainable Artificial Intelligence (XAI) is focused on Explainable Deep Learning (XDL) (cf. \cite{10.1613/jair.1.13200}). 

The taxonomy of the trends identified for explainability techniques related to Deep Learning Models (DLMs) distinguishes, among others, model-agnostic techniques (MATs) and model-specific techniques (MSTs) (cf. \cite{BARREDOARRIETA202082}). An explainer in MATs is capable of explaining any model (cf. the LIME technique \cite{10.1145/2939672.2939778}). An explainer in MSTs is correlated with a given deep learning model (cf. the DeepLIFT technique \cite{shrikumar2019}).

\begin{figure}[!bht] 
\centering
\includegraphics[width=0.95\textwidth]{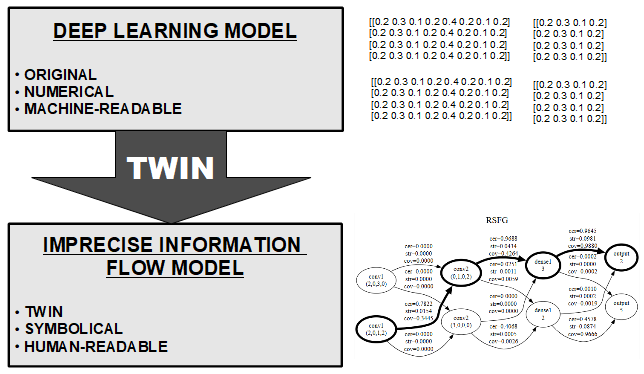}
\caption{An idea of creation of a readable twin (an imprecise information flow model) for an unreadable original model (a deep learning model).}
\label{Fig:idea}
\end{figure}

In our research, we are developing a new technique that can be classified as MST. This technique is provided with an acronym HuReTEx (Human Readable Twin Explainer for Deep Learning Models). An idea of transformation of a deep learning model (DLM) into imprecise information flow model (IIFM) via a Sequential Information System (SIS) is shown in Figure \ref{Fig:idea}. In this transformation, DLM is original, numerical and machine-readable, while IIFM is a twin of DLM that is symbolical and human-readable. 

\begin{figure}[!bht] 
\centering
\includegraphics[width=0.95\textwidth]{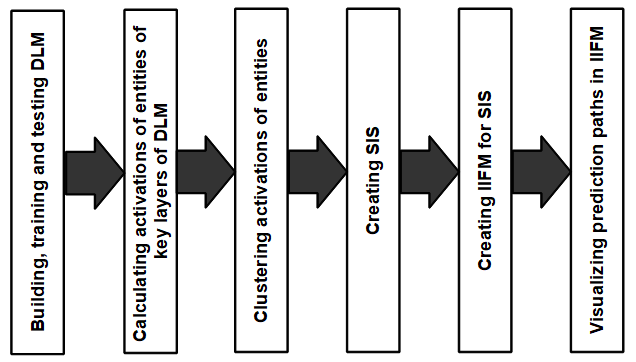}
\caption{A flowchart of the transformation procedure of DLM into IIFM.}
\label{Fig:flowchart}
\end{figure} 

HuReTEx can be treated as a reference to the ideas of mirror worlds (cf. \cite{Gelernter1991}), as well as digital twins (cf. \cite{LIU2021346}), as it is shown in Table \ref{tab:digital_and_readable}. Moreover, the proposed approach refers to twin-systems for XAI (cf. \cite{ijcai2019p376}). For a model that is unreadable to humans, its readable twin is built.  
\begin{table}[!bht]
    \centering
    \begin{tabular}{|c|c|}
    \hline 
    Model & Description \\
    \hline \hline
        Digital twin model & A combination of a physical object \\
        & and its digital representation in virtual space. \\
        \hline
        Readable twin model &  A combination of an unreadable deep learning model \\
        & and its readable representation. \\
        \hline
    \end{tabular}
    \caption{Digital twin model vs. readable twin model.}
    \label{tab:digital_and_readable}
\end{table}

Transformation is preformed in several main stages (see a flowchart in Figure \ref{Fig:flowchart}). Rough set flow graphs (RSFGs) \cite{Pawlak_TRS_III} (used as IIFMs) and triangular norms or co-norms together with evolutionary algorithms can be used to mine and visualize the most confident prediction paths in RSFGs explaining decisions proposed by DLMs (see next section).   

In our approach, the twin model is built on the basis of aggregated artifacts (at specific levels of abstraction) generated by individual model layers for training data. In this case, the explanations given by the model are not generated for individual input cases only.

\section{From Unreadable Models to Their Readable Twins}

In this section, the consecutive steps of the procedure to create a readable twin of an unreadable deep learning model are described. A readable twin model has the form of a rough set flow graph (RSFG). Elements of RSFG correspond to the original deep learning model (DLM) as it is shown in Table \ref{tab:DLM_RSFG}.

\begin{table}[!bht]
    \centering
    \begin{tabular}{|c|c|}
    \hline 
    Deep Learning Model (DLM) & Rough Set Flow Graph (RSFG) \\
    \hline \hline
        A convolutional layer of DLM & A node layer of RSFG \\
        \hline
        Clusters of artifacts generated by filters & Nodes in a layer of RSFG \\
        of a convolutional layer of DLM & \\
        \hline
        A dense layer of DLM & A node layer of RSFG \\
        \hline
        Clusters of artifacts generated by neurons & Nodes in a layer of RSFG \\
        of a dense layer of DLM & \\
        \hline
    \end{tabular}
    \caption{DLM vs. RSFG.}
    \label{tab:DLM_RSFG}
\end{table}

It is worth noting that we have omitted the input layer and the  flatten layer because they are not trained in deep learning models and therefore they do not acquire knowledge that could be used in the explanation process.

\begin{figure}[!bht] 
\centering
\includegraphics[width=0.8\textwidth]{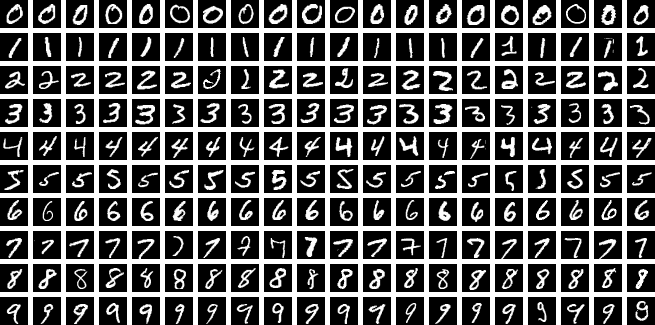}
\caption{A fragment of the MNIST data set (by Suvanjanprasai - Own work, CC BY-SA 4.0 https://commons.wikimedia.org/w/index.php?curid=156115980).}
\label{Fig:mnist}
\end{figure}

An illustrative example is presented for a simple deep learning model built using the Keras library \cite{chollet2015keras} to classify images from the well-known MNIST database of handwritten digits (see Figure \ref{Fig:mnist}) \cite{deng2012mnist}. The training data set includes 48000 images. 

\subsubsection{Step 1:} 
\begin{figure}[!bht] 
\centering
\includegraphics[width=0.9\textwidth]{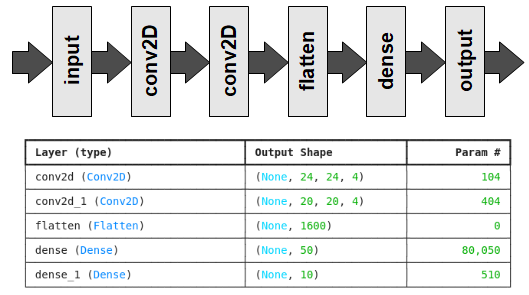}
\caption{Creation of a readable twin for an unreadable original model: Step 1.}
\label{Fig:step1}
\end{figure}

Building, training and testing a deep learning model (DLM) on training cases (images). The model can be  built (see Figure\ref{Fig:step1}), trained and tested in a standard way. 

\subsubsection{Step 2:} 
\begin{figure}[!bht] 
\centering
\includegraphics[width=0.9\textwidth]{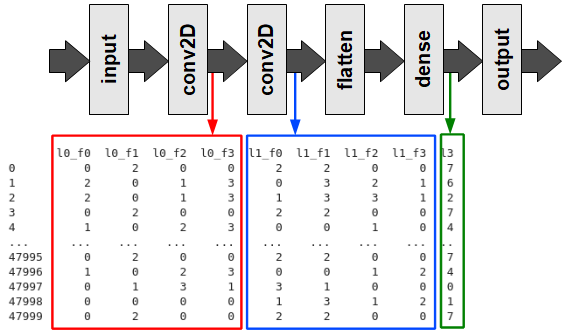}
\caption{Creation of a readable twin for an unreadable original model: Step 2.}
\label{Fig:step2}
\end{figure}

Calculating and clustering activations generated by key layers of DLM for training cases (images). After training and testing DLM on images from the MNIST data set, the activations of the model entities for each image from the training data set are calculated and clustered. A good choice may be agglomerative clustering leading to a hierarchical structure of clusters, which can be easily mapped to a hierarchical conceptual knowledge structure about artifacts. In an example in Figure \ref{Fig:step2}, the cluster identifiers, to which artifacts generated for images from the training data set belong, are shown.

\subsubsection{Step 3:} 
\begin{figure}[!bht] 
\centering
\includegraphics[width=0.9\textwidth]{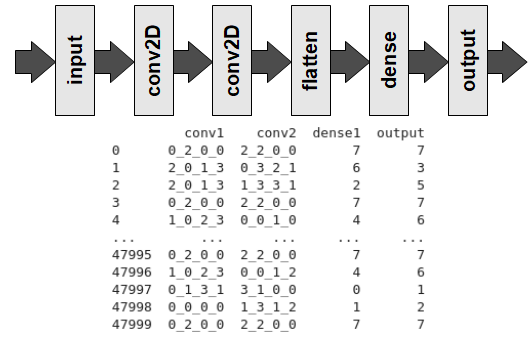}
\caption{Creation of a readable twin for an unreadable original model: Step 3.}
\label{Fig:step3}
\end{figure}

Creating a sequential information system (SIS). The underlying information for generation of a rough set flow graph is arranged in the form of a sequential information system (see an example in Figure \ref{Fig:step3}), that is, an information system (called in \cite{Pawlak1991} a knowledge representation system) with an ordered set of attributes (presented in columns). For convolutional layers, attribute values are tuples of the identifiers of clusters obtained in Step 2. For dense layers, attribute values are identifiers of clusters obtained in Step 2. For an output layer, attribute values are labels assigned to training cases (images). 

\subsubsection{Step 4:} 
\begin{figure}[!bht] 
\centering
\includegraphics[width=0.9\textwidth]{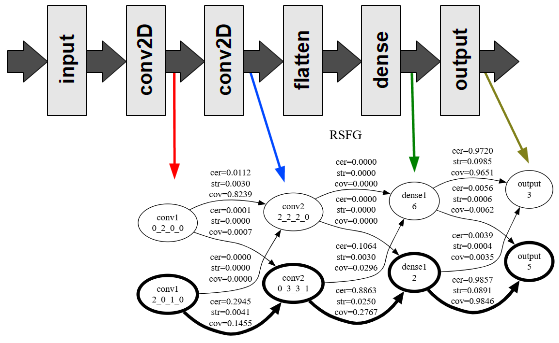}
\caption{Creation of a readable twin for an unreadable original model: Step 4.}
\label{Fig:step4}
\end{figure}

Creating a rough set flow graph (RSFG). RSFG has a layered structure. Each layer corresponds to one attribute (column) of SIS obtained in Step 3. The nodes in a given layer represent the values of a given attribute of SIS. An imprecise information flow between nodes is described by three coefficients (certainty, strength, and covering) assigned to edges connecting nodes. The certainty of a given edge connecting nodes $n$ and $n'$ determines how many times a transition occurs from node $n$ to node $n'$ relative to all transitions from node $n$ to other nodes. The covering of a given edge connecting nodes $n$ and $n'$ determines how many times there is a transition from node $n$ to node $n'$ relative to all transitions to node $n'$ from other nodes. The strength of a given edge connecting nodes $n$ and $n'$ determines how many times there is a transition from node $n$ to node $n'$ out of all transitions between the layer containing $n$ and the layer containing $n'$. A sample fragment of RSFG is shown in Figure \ref{Fig:step4}. 

\subsubsection{Step 5:} 
\begin{figure}[!bht] 
\centering
\includegraphics[width=0.9\textwidth]{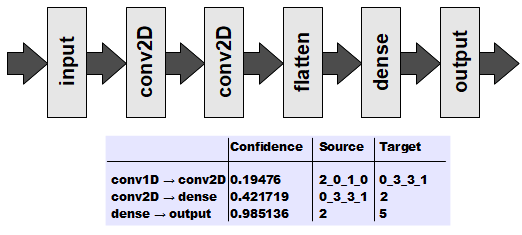}
\caption{Creation of a readable twin for an unreadable original model: Step 5.}
\label{Fig:step5}
\end{figure}

Determining confident prediction paths in RSFG obtained in Step 4. In general, rough set flow graphs can represent a large number of paths (sequences of edges) between nodes. In the presented approach, we propose to use the evolutionary algorithm (EA) to mine the most important paths. In EA, chromosomes are sequences of nodes in consecutive layers of RSFG. The fitness function is defined on the basis of confidence (a harmonic mean of certainty and covering) of edges between nodes in sequences represented by chromosomes. The goal is to find the sequences of nodes with the highest possible aggregated confidence value. To aggregate confidences of individual edges between nodes in sequences, triangular norms or co-norms \cite{Klement2000} are used. A sample of the confident prediction path found by EA is shown in Figure \ref{Fig:step5}. 

\subsubsection{Step 6:} 
\begin{figure}[!bht] 
\centering
\includegraphics[width=0.9\textwidth]{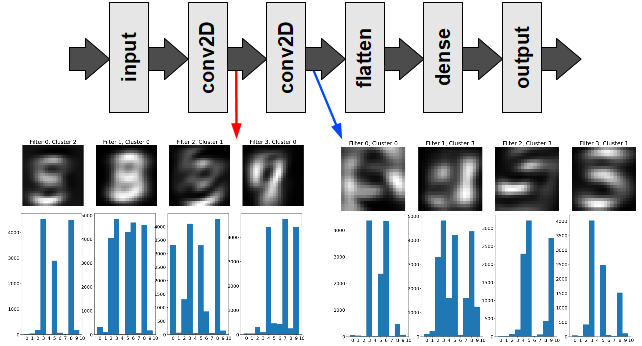}
\caption{Creation of a readable twin for an unreadable original model: Step 6.}
\label{Fig:step6}
\end{figure}

Visualizing confident prediction paths. Visualization of confident prediction paths is the quintessence of the presented approach. A sample of the confident prediction path visualization is shown in Figure \ref{Fig:step6}. Aggregated artifacts generated by the filters were associated with histograms showing how many cases (images) of given classes generated artifacts assigned to a given cluster.

\begin{figure}[!bht] 
\centering
\includegraphics[width=0.95\textwidth]{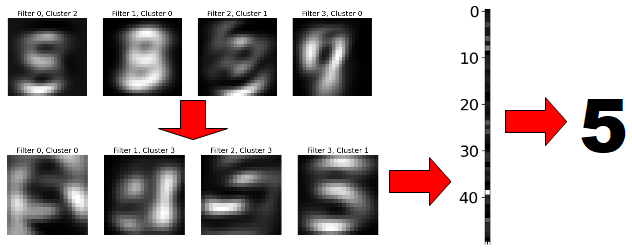}
\caption{Visual summarization of one of the most confident paths.}
\label{Fig:summarizarion}
\end{figure}

Visual summarization of one of the most confident paths is shown in Figure \ref{Fig:summarizarion}.

\section{Conclusions}

We have shown how to use rough set flow graphs to model the information flow in sequential deep learning models. 
The rough set flow graph model explains what the model has learned. 
This can be considered as a clear representation of the knowledge acquired by the model, as opposed to the knowledge hidden in the weights and coefficients of the original deep learning model. 
Thus, the idea of creating readable twins of unreadable models has been presented.

In further research, we propose to incorporate ontologies of artifacts generated by model layers. Thanks to this, it will be possible to implement the following path: hierarchical clustering of artifacts, hierarchical structure of concepts describing clusters, and conceptual knowledge structure of artifacts. 
The use of other graphical models will enable us to visualize the prediction processes performed by the model. The next step on this path could be the description of artifacts in natural language. It is also planned to use other information flow models, such as high-level Petri nets, for example, Petri nets over ontological graphs \cite{Pancerz_PP-RAI_2024}.

\bibliographystyle{splncs04}
\bibliography{RTofUM}

\begin{thebibliography}{10}
\providecommand{\url}[1]{\texttt{#1}}
\providecommand{\urlprefix}{URL }
\providecommand{\doi}[1]{https://doi.org/#1}

\bibitem{BARREDOARRIETA202082}
{Barredo Arrieta}, A., Díaz-Rodríguez, N., {Del Ser}, J., Bennetot, A.,
  Tabik, S., Barbado, A., Garcia, S., Gil-Lopez, S., Molina, D., Benjamins, R.,
  Chatila, R., Herrera, F.: Explainable artificial intelligence {(XAI)}:
  Concepts, taxonomies, opportunities and challenges toward responsible {AI}.
  Information Fusion  \textbf{58},  82--115 (2020).
  \doi{10.1016/j.inffus.2019.12.012}

\bibitem{chollet2015keras}
Chollet, F., et~al.: Keras. \url{https://keras.io} (2015)

\bibitem{deng2012mnist}
Deng, L.: The {MNIST} database of handwritten digit images for machine learning
  research. IEEE Signal Processing Magazine  \textbf{29}(6),  141--142 (2012).
  \doi{10.1109/MSP.2012.2211477}

\bibitem{Gelernter1991}
Gelernter, D.: Mirror Worlds: Or: The Day Software Puts the Universe in a
  Shoebox...How It Will Happen and What It Will Mean. Oxford University Press
  (1991). \doi{10.1093/oso/9780195068122.001.0001}

\bibitem{ijcai2019p376}
Kenny, E.M., Keane, M.T.: {Twin-Systems to Explain Artificial Neural Networks
  using Case-Based Reasoning: Comparative Tests of Feature-Weighting Methods in
  ANN-CBR Twins for XAI}. In: Proceedings of the Twenty-Eighth International
  Joint Conference on Artificial Intelligence ({IJCAI-19}). pp. 2708--2715.
  International Joint Conferences on Artificial Intelligence Organization
  (2019). \doi{10.24963/ijcai.2019/376}

\bibitem{Klement2000}
Klement, E.P., Mesiar, R., Pap, E.: Triangular Norms. Springer, Dordrecht
  (2000)

\bibitem{LIU2021346}
Liu, M., Fang, S., Dong, H., Xu, C.: Review of digital twin about concepts,
  technologies, and industrial applications. Journal of Manufacturing Systems
  \textbf{58},  346--361 (2021). \doi{10.1016/j.jmsy.2020.06.017}

\bibitem{Pancerz_PP-RAI_2024}
Pancerz, K.: A {Python} toolkit for dealing with {Petri} nets over ontological
  graphs. \url{https://arxiv.org/abs/2504.08006} (2025).
  \doi{10.48550/arXiv.2504.08006}

\bibitem{Pawlak1991}
Pawlak, Z.: Rough Sets. Theoretical Aspects of Reasoning about Data. Kluwer
  Academic Publishers, Dordrecht (1991)

\bibitem{Pawlak_TRS_III}
Pawlak, Z.: Flow graphs and data mining. In: Peters, J.F., Skowron, A. (eds.)
  Transactions on Rough Sets III, pp. 1--36. Springer-Verlag, Berlin Heidelberg
  (2005). \doi{10.1007/11427834\_1}

\bibitem{10.1613/jair.1.13200}
Ras, G., Xie, N., van Gerven, M., Doran, D.: Explainable deep learning: A field
  guide for the uninitiated. J. Artif. Int. Res.  \textbf{73} (2022).
  \doi{10.1613/jair.1.13200}

\bibitem{10.1145/2939672.2939778}
Ribeiro, M.T., Singh, S., Guestrin, C.: {"Why Should I Trust You?"}: Explaining
  the predictions of any classifier. In: Proceedings of the 22nd ACM SIGKDD.
  pp. 1135--1144. KDD '16 (2016). \doi{10.1145/2939672.2939778}

\bibitem{shrikumar2019}
Shrikumar, A., Greenside, P., Kundaje, A.: Learning important features through
  propagating activation differences. \url{https://arxiv.org/abs/1704.02685}
  (2019). \doi{10.48550/arXiv.1704.02685}

\end{thebibliography}

\end{document}